\def\year{2017}\relax
\documentclass[letterpaper]{article}
\usepackage{aaai17}
\usepackage{times}
\usepackage{helvet}
\usepackage{courier}
\usepackage{natbib}
\usepackage{graphicx}
\begin{document}
\title{Latent Tree Analysis}
\author{Nevin L. Zhang\\
Department of Computer Science \& Engineering \mbox{   } \\
The Hong Kong University of Science \& Technology\\
{\tt lzhang@cse.ust.hk}\\
\And
Leonard K. M. Poon\\
\mbox{   } Department of Mathematics \& Information Technology\\
The Education University of Hong Kong \\
{\tt kmpoon@eduhk.hk}
}
\maketitle
\begin{abstract}
Latent tree analysis seeks to model the correlations among a set of random variables using a tree of latent variables. It was proposed as an improvement to latent class analysis ---  a method widely used in social sciences and medicine to identify homogeneous subgroups in a population. It provides new and fruitful perspectives on a number of machine learning areas, including cluster analysis, topic detection, and deep probabilistic modeling. This paper gives an overview of the research on latent tree analysis and various ways it is used in practice.

\end{abstract}

Much of machine learning is about modeling and utilizing correlations among variables. In classification, the task is to establish relationships between attributes and class variables so that unseen data can be classified accurately. In Bayesian networks, dependencies among variables are  represented as directed acyclic graphs and the graphs are used to facilitate efficient probabilistic inference. In topic models, word co-occurrences are accounted for by assuming that all words are generated   probabilistically from the same set of topics, and the generation process is reverted via statistical inference to determine the topics. In deep belief networks, correlations among observed units are modeled using multiple levels of hidden units, and the top-level hidden units are used as a representation of the data for further analysis.

{\em Latent tree analysis (LTA)} seeks to model the correlations among a set of observed variables using a tree model, called {\em latent tree model (LTM)}, where the leaf nodes represent observed variables and the internal nodes represent latent variables.The dependence between two observed variables is explained by the path between them.

Despite their simplicity, LTMs subsume two classes of models widely used in academic research. The first one is {\em latent class models (LCMs)} \citep{lazarsfeld1968, knott1999latent}, which are LTMs with a single latent variable. They are used for categorical data clustering in social sciences and medicine. The second class is probabilistic phylogenetic trees \citep{durbin98bio}, which are a tool for determining the evolution history of a set of species. Phylogenetic trees are special LTMs where the model structures are binary (bifurcating) trees and all the variables have the same  number of possible states.

LTA  also provides new and fruitful perspectives on a number of machine learning areas. One area is cluster analysis. Here finite mixture models such as LCMs are commonly used. A finite mixture model has one latent variable and consequently it gives one soft partition of data. An LTM typically has multiple latent variables and hence LTA yields multiple soft partitions of data simultaneously. In other words, LTA performs {\em multidimensional clustering} \citep{chen12model, liu2013greedy}. It is interesting because complex data usually have multiple facets and can be meaningfully clustered in multiple ways.

Another area is topic detection. Applying LTA to text data, we can partition a collection of documents in multiple ways. The document clusters in the partitions can be interpreted as topics. Furthermore, it is possible to learn {\em hierarchical LTMs} where the latent variables are organized into multiple layers. This leads to an alternative method for hierarchical topic detection \citep{liu14hierarchical, chen2015progressive}, which has been shown to find more meaningful topics and topic hierarchies than the state-of-the-art method based on latent Dirichlet allocation \citep{paisley2015nested}.

The third area is deep probabilistic modeling. Hierarchical LTM and deep belief network (DBN) \citep{hinton2006fast} are similar in that they both consist of multiple layers of variables, with an observed layer at the bottom and multiple layers of hidden units on top of it. One difference is that, in DBN, units from adjacent layers are fully connected, while HLTM is tree-structured.  It would be interesting to explore the middle ground between the two extreme and develop algorithms for learning what might be called sparse DBNs. Learning structures for deep models is an interesting open problem. Extension of LTA might offer one solution  \citep{chen2016sparse}.

The concept of latent tree models was introduced in~\citep{zhang02hierarchical,zhang04hierarchical}, where they were referred to as hierarchical latent class models. The term ``latent tree models" first appeared in~\citep{zhang08latent,wang08alatent}.  \cite{mourad2013survey} surveyed the research on latent tree models as of 2012 in details. This paper provides a concise overview of the methodology. The exposition are more conceptual and less technical than \citep{mourad2013survey}. Developments after 2012 are also included.

\section{Preliminaries}
A {\em latent tree model (LTM)} is a tree-structured Bayesian network \citep{pear1988probabilistic}, where the leaf nodes represent observed variables and the internal nodes represent latent variables. An example is shown in Figure 1 (a). All variables are assumed to be discrete. The model parameters include a marginal distribution for the root $Y_1$ and a conditional distribution for each of the other nodes given its parent. The product of the distributions defines a joint distribution over all the variables.

By changing the root from $Y_1$ to $Y_2$ in Figure 1 (a), we get another model shown in  (b). The two models are {\em equivalent} in the sense that they represent the same set of distributions over the observed variables $X_1$, \ldots, $X_5$ \citep{zhang04hierarchical}. It is not possible to distinguish between equivalent models based on data. This implies that edge orientations in LTMs are unidentifiable. It therefore makes more sense to talk about undirected LTMs, which is what we do in this paper.  One example is shown in Figure 1 (c). It represents an equivalent class of directed models, which includes the two models shown in (a) and (b) as members. In implementation, an undirected model is represented using an arbitrary directed model in the equivalence class it represents.

In the literature, there are variations of LTMs where some internal nodes are observed \citep{choi11learning} and/or the variables are continuous \citep{poon2010variable,kirshner2012latent,song2014nonparametric}. In this paper, we focus on basic LTMs as defined in the previous two paragraphs.

We use $|W|$ to denote  the number of possible states of a variable $W$.
An LTM is {\em regular} if, for any latent node $Z$, we have that
\(|Z| \leq \frac{\prod_{i=1}^k|Z_i|}{\max_{i=1}^k|Z_i|},\)
\noindent where $Z_1$, \ldots, $Z_k$ are the neighbors of $Z$, and that the inequality holds strictly when $k=2$. For any irregular LTM, there is a regular model that has fewer parameters and represents that same set of distributions over the observed variables \citep{zhang04hierarchical}. Consequently, we focus only on regular models.

\begin{figure}[b]
\includegraphics[width=8.5cm]{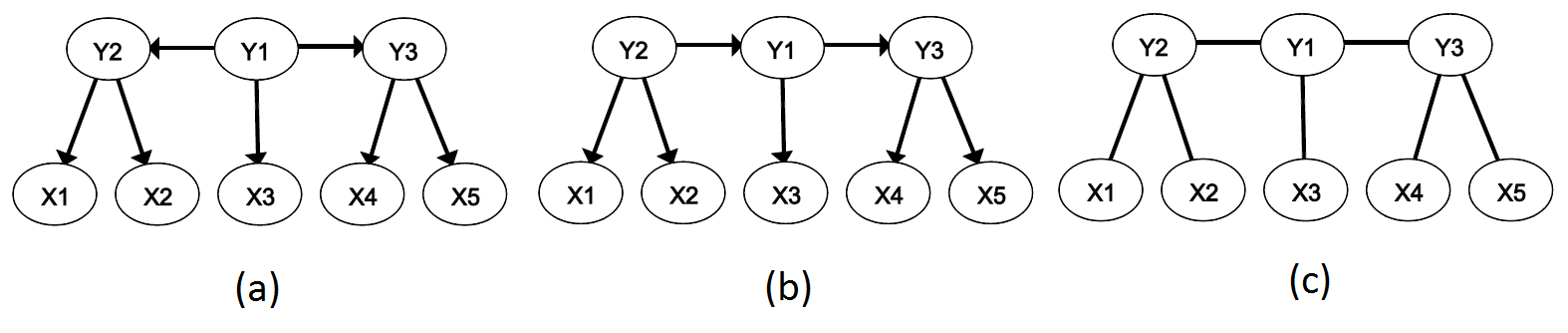}
\vspace{-0.5cm}
\caption{\small The undirected latent tree model in (c) represents an equivalent class of directed latent tree models, which includes (a) and (b) as members.}
\vspace{-0.5cm}
\end{figure}

\section{Learning Latent Tree Models}
To fit an LTM to a dataset, one needs to determine: (1) the number of latent variables, (2) the number of possible states for each latent variable, (3) the connections among all the variables, and (4) the probability distributions.

There are three commonly used algorithms for learning LTMs: EAST (Expansion, Adjustment and Simplification until Termination) \citep{chen12model}, BI (Bridged Islands) \citep{liu2013greedy}, and CLRG (Chow-Liu and Recursive Grouping) \citep{choi11learning}. EAST is a search-based algorithm and it aims to find the model with the highest BIC score. It is the slowest among the three and finds better models, as measured by held-out likelihood, than the other two algorithms \citep{liu2013greedy}. It is often used to analyze survey data from medicine and social sciences, which typically contain dozens of observed variables.

BI first divides the observed variables into unidimensional subsets. A set of variables is {\em unidimensional} if the correlations among them can be properly modeled using a single latent variable, which is determined using a test that compares the best one-latent-variable model and the best two-latent-variable. BI then introduces a latent variable for each unidimensional subset to form a LCM. The LCMs are metaphorically called islands. The latent variables in the islands are linked up using Chow-Liu's algorithm \citep{chow1968approximating} to form a global model.

CLRG first constructs a tree over the observed variables using Chow-Liu's algorithm. It then recursively transforms patches of the model by adding latent variables and/or re-arranging the edges. Here a patch consists of an internal node and its neighbors. {\em Information distances}, as defined in \citep{erdos1999few}, between pairs of variables in the patch are estimated from data. They are used to transform the patch into a latent tree based on a theorem which states that, if the variables are indeed from a tree model, information distances between them are additive w.r.t.\ the tree.

Empirical results reported in \citep{liu2013greedy} show that BI consistently yields better models than CLRG in terms of held-out likelihood. BI scales up well if progressive EM is used to estimate the parameters of the intermediate models, and was able to handle a text dataset with  10,000 distinct words (variables) and 300,000 documents in around 11 hours on a single machine \citep{chen2015progressive}. CLRGC also scales up well if parameter learning is based on the method of moments and tensor decomposition, and was able to process a medical dataset with around 1,000 diagnosis categories (variables) and 1.6 million patient records in around 4.5 hours on a single machine \citep{huang2015scalable}.

\section{Improving Latent Class Analysis}

As mentioned before, a {\em latent class model (LCM)} is an LTM with a single latent variable, and LCA refers to the process of fitting an LCM to a dataset. The latent variable gives a soft partition of data and its states represent clusters in the partition. LCA is hence a technique for cluster analysis and it is widely used in social, behavioral and health sciences \citep{collins2010latent}. In medical research, it is used to identify subtypes of diseases, for instance major depression, where good standards are not available \citep{van2013latent,li2014subtypes}.

A major issue with LCA is the assumption that all the observed variables are mutually independent given the latent variable. In other words, the observed variables are assumed to be independent in each cluster of data. The assumption is hence  known as the {\em local independence assumption}. It is easily violated in practice and casts doubts on the validity of the clustering results \citep{van2013latent}.

 LTMs provide a natural framework where the local independence assumption can be relaxed.
We illustrate the point first using an example from \citep{fu2016identification}, where the task is to divide a collection of patients with vascular mild cognitive impairment (VMCI) into subclasses based on 27 symptoms that are related to the concept of Qi Deficiency in traditional Chinese medicine. Figure 2 (a) shows the LCM learned for the task and (b) shows the LTM. In the LTM, intermediate latent variables are introduced between the observed variables at the bottom and the clustering variable $Z$  at the top. We refer to such models as {\em LTM-based unidimensional clustering models}. The local independence assumption is relaxed because, given the clustering variable $Z$, the observed variables are no longer mutually independent.

 The relaxation of the local independence leads to better model fit. As a matter of fact, the BIC score of the LTM is -10,022, which is higher than that of the LCM, which is -10,164. It also leads better clustering results as will be explained later in this section. In addition, the LTM is also intuitively more reasonable than the LCM.  For example, the symptoms ``dry stool or constipation" and ``asthenia of defecation" are both about difficulties with defecation. It is hence reasonable to connect them to  $Z$ via an intermediate variable ($Y_{01}$), which can be interpreted as the impact of Qi Deficiency on defecation.
 %The symptoms ``palpitation",``chest oppression" and ``short of breath" are, according to TCM theory, manifestations of Qi Deficiency in the heart.  It is hence reasonable to connect them to the clustering variable via a intermediate variable $W$.
 %In addition, the LTM divides the patients into three clearly meaningful clusters: no Qi Deficiency, Qi Deficiency related to the heart, and Qi Deficiency related to the kidney.
%
%\cite{fu2016identification} obtained the model shown in Figure 2 (b)  as follows. The whole data set with 93 observed variables was first analyzed using the EAST algorithm to produce a global model, which was then used to determine the intermediate latent variables in (2). If two observed variables are directly connected to the same latent variables in the global model, then they were connected to the clustering variable $Z$ in (b) via an intermediate latent variables. A search process was carried out to check with the intermediate variables and the observed variables directed connected to $Z$ should be merged.  The variable $W$ was introduced by the search process.

\begin{figure}[t]
\includegraphics[width=8.5cm]{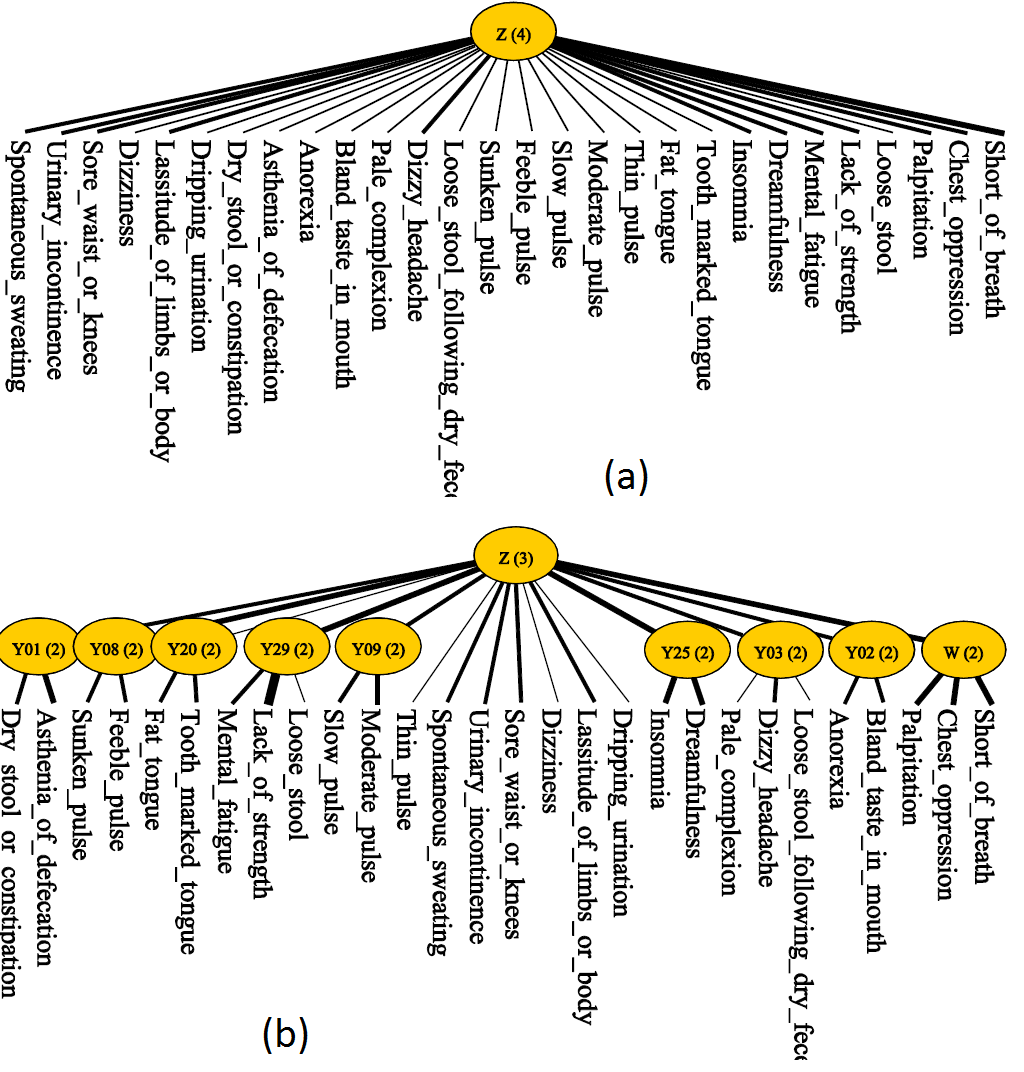}
\vspace{-0.5cm}
\caption{\small Models produced on a  dataset by LCA (a) and LTA (b). Numbers next to latent variables are the numbers of possible states. Edge widths indicate mutual information between variables.}
\vspace{-0.5cm}
\end{figure}

 LTM-based clustering models  can be obtained by first fitting an LTM to data using EAST or BI. This divides the observed variables into groups, called {\em sibling clusters}, each consisting of the variables directly connected to a latent variable. All sibling clusters are unidimensional and the correlations among the members are properly modeled by corresponding latent variables. The latent variables, with one possible exception\footnote{This is determined by search \citep{liu2015unidimensional}. If a latent variable is not used as a feature, then all the observed variables in its sibling cluster are.},  are then used as features for clustering instead of the individual observed variables, resulting in a model similar to the one shown in Figure 2 (b).

A standard way to evaluate a clustering algorithm is to start with a labeled dataset, remove the class labels,  run the  algorithm to partition the resulting unlabeled dataset, and measure the performance using mutual information between the partition obtained with the partition induced by the class labels. Following this practice,
\cite{liu2015unidimensional} have compared LTM-based cluster analysis with LCA on 30 datasets from UCI, and found that the LTM-based method
   outperforms  LCA in most cases and often significantly.

\section{Multidimensional Clustering}

Complex data usually have multiple facets and can be meaningfully partitioned in multiple ways. For example, a student population can be clustered in one way based on academic performances and another based on extracurricular activities. Movie reviews can be clustered based on  sentiment (positive or negative) or genre (comedy,
action, war, etc.). The respondents in a social survey can be clustered based on
demographic information or views on social issues.

 There are efforts on developing clustering algorithms that produce multiple partitions of data, with each partition being based, solely or primarily,  on a different subset of attributes. We call them {\em multidimensional clustering}  methods. (This is not to be confused with multi-view clustering, which combines information from different views of data to improve the quality of one single partition.)
 There are sequential methods that aim at obtaining additional partitions of data that are novel w.r.t a previous partition, which can, for instance, be obtained using K-means \citep{cui2007non,gondek2007non,qi2009principled,bae2006coala}. There are also  methods that produce multiple partitions simulataneously \citep{jain2008simultaneous,niu2010multiple}. Those methods try to optimize the quality of each individual partition while keeping different partitions as dissimilar as possible. All of the methods are limited in the number of different partitions they produce, which is typically 2.

An LTM typically has multiple latent variables, and each of them can be interpreted as representing one soft partition of data. As such, LTA is a natural tool for multidimensional clustering \citep{chen12model,liu2013greedy}. LTA can determine the number of partitions automatically and it is not limited in the number of partitions.

We illustrate the use of LTA for multidimensional clustering using an example from
\citep{chen12model}, where a survey dataset from ICAC --- Hong Kong's anti-corruption agency --- was analyzed using the EAST algorithm. The structure of the resulting model is shown in Figure 3.  There are 9 latent variables. It is clear that the latent variable $Y_2$ represents a partition of the respondents primarily based on demographic information; $Y_3$  represents a partition based on people's tolerance toward corruption; $Y_4$  represents a partition based on people's view on ICAC's performance; $Y_6$  represents a partition based on people's view on the level of corruption; $Y_5$  represents a partition based on people's view on the trend of corruption; and so on. It makes sense that there are direct dependencies between $Y_2$ (demographic information) and $Y_3$ (tolerance toward corruption); between $Y_4$ (ICAC performance)  and $Y_6$ (corruption level); and between  $Y_4$ (ICAC performance)  and $Y_5$ (corruption trend).

The conditional distributions of two observed variables given $Y_3$ are  given in Figure 3. The states of the observed variables are $s_0$ (totally intolerable), $s_1$ (intolerable), $s_2$ (tolerable), and $s_3$ (totally tolerable).  Based on the  distributions, the three states of $Y_3$ are interpreted as classes of people who find corruption totally intolerable ($Y_3=s_0$), intolerable ($Y_3=s_1$), and tolerable ($Y_3=s_2$) respectively. The distributions also suggest that people who are tough on corruption ($Y_3=s_0$) are equally tough toward corruption in the government and corruption in
the business sector, while people who are lenient towards corruption are more lenient toward corruption in the business
sector than corruption in the government.

\begin{figure}[t]
\includegraphics[width=9.0cm]{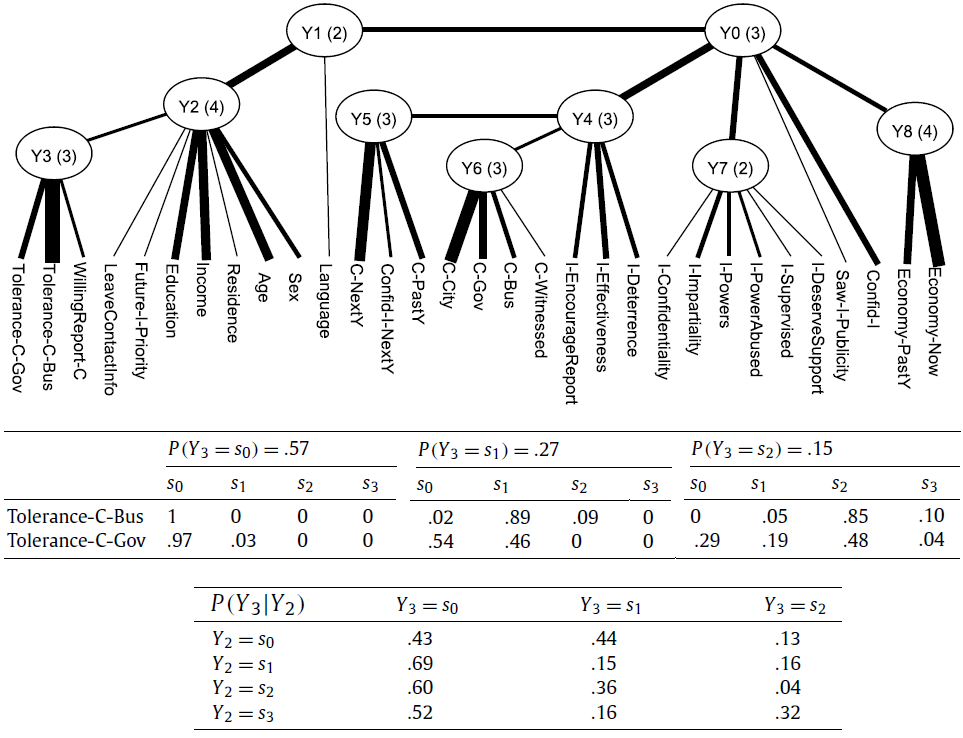}
\vspace{-0.5cm}
\caption{\small The structure of the LTM obtained from the ICAC data and some of the probability distributions. Abbreviations: C -- Corruption, I -- ICAC, Y -- Year, Gov -- Government, Bus -- Business Sector.
Meanings of variables: Tolerance-C-Gov -- `tolerance towards corruption in the government'; C-City -- `level of corruption in the city';
C-NextY --  `change in the level of corruption next year'; I-Effectiveness --  `effectiveness of ICAC's work'; I-Powers --  `ICAC powers'; Confid-I
--  `confidence in ICAC'; etc.}
\vspace{-0.5cm}
\end{figure}

Based on the conditional distributions of the demographic variables given $Y_2$, the four states of the latent variable are interpreted as:  $Y_2=s_0$ -- low income youngsters; $Y_2=s_1$ -- women with no/low income; $Y_2=s_2$ -- people with good education and good income; $Y_2=s_3$ -- people with poor education and average income.
The conditional distribution $P(Y_3|Y_2)$ is also given in Figure 3. It suggests that
 people with good education and good income ($Y_2=s_2$) are the toughest toward corruption, while people with poor education and average income  ( $Y_2=s_3$) are the most lenient. This is intuitively appealing and can be a
hypothesis for social scientist to verify further.

\cite{liu2013greedy} quantitatively compared LTA with the alternative methods for multidimensional clustering mentioned above on the WebKB dataset, which has two ground-truth partitions (class labels). The class labels were first removed and all algorithms were run to recover the ground-truth partitions. LTA significantly outperforms all the alternative methods.

\section{Hierarchical Topic Detection}

\begin{figure}[t]
\includegraphics[width=8.5cm]{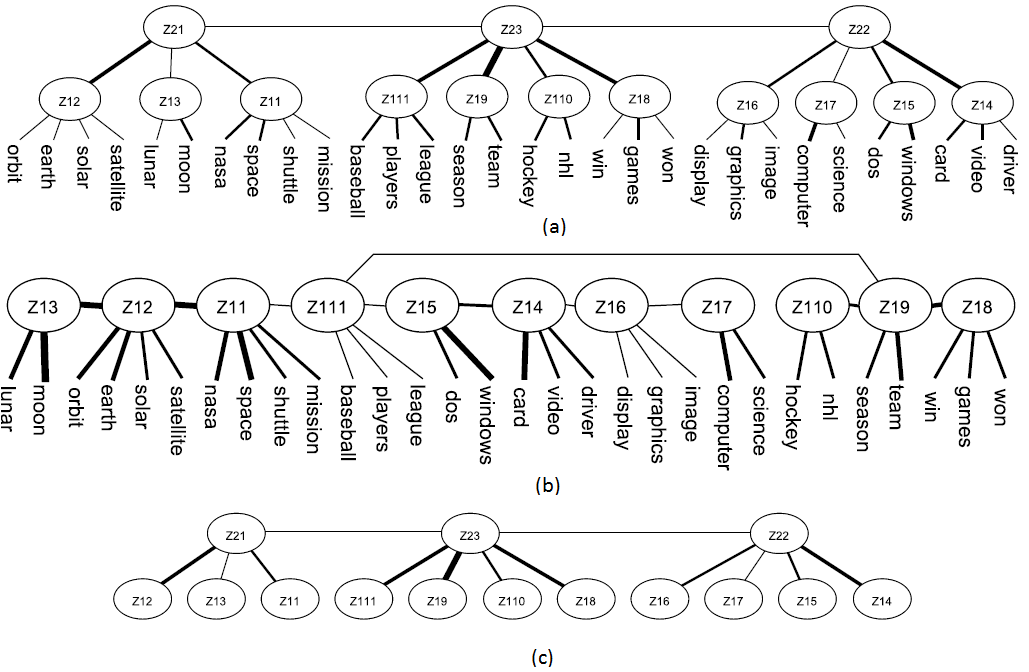}
\vspace{-0.5cm}
\caption{\small  Hierarchical model obtained on a toy text dataset by HLTA (a) and intermediate models (b, c) created by HLTA.}
\vspace{-0.5cm}
\end{figure}

\cite{liu14hierarchical} propose a method to analyze text data and obtain models such as the one shown in Figure 4 (a). There is a layer of observed variables at the bottom, and multiple layers of latent variables on top. Such models are called {\em hierarchical latent tree models (HLTMs)} and the process of learning HLTMs is called {\em hierarchical latent tree analysis (HLTA)}.

The observed variables are binary variables that represent the absence/presence of words in documents. The level-1 latent variables model patterns of probabilistic word co-occurrence, and latent variables at  higher levels model co-occurrences of patterns at the level below. For example, $Z_{14}$ captures the co-occurrence of the words ``card", ``video" and ``driver"; $Z_{15}$ captures the co-occurrence ``windows" and ``dos"; $Z_{22}$ captures the co-occurrence of the patterns represented by $Z_{14}$, $Z_{15}$, $Z_{16}$ and $Z_{17}$; and so on.

The latent variables are also binary. Each of them partitions the documents into two clusters. Information about some of the partitions is given below.

\vspace{0.3cm}

{\small

\begin{tabular}{cc}

\begin{tabular}{c}
\hspace{-1cm}
\begin{tabular}{lll}
\hline
& $s_0$  & $s_1$  \\
$Z_{14}$ &  (0.88) & (0.12)  \\ \hline
 card   & 0.01 & 0.47 \\
video  & 0.02 & 0.32 \\
driver & 0.03 & 0.20
\\ \hline
\end{tabular}
 \\

\\
\hspace{-1cm} \begin{tabular}{lll}
\hline
& $s_0$  & $s_1$  \\
$Z_{15}$ &  (0.85) & (0.15)  \\ \hline
windows   & 0.01 & 0.67 \\
dos  & 0.01 & 0.30 \\
     & &
\\ \hline
\end{tabular}

\end{tabular}

&
 \begin{tabular}{lll}
\hline
& $s_0$  & $s_1$  \\
$Z_{22}$ &  (0.76) & (0.24)  \\ \hline
windows     & 0.04 & 0.34 \\
card        & 0.02 & 0.22 \\
graphics    & 0.02 & 0.17 \\
video       & 0.01 & 0.15 \\
dos         & 0.02 & 0.16 \\
computer    & 0.05 & 0.21 \\
display     & 0.02 & 0.10 \\
drive       & 0.02 & 0.10
\\ \hline
\end{tabular}

\end{tabular}

}

\vspace{0.3cm}

We see that the two clusters given by $Z_{14}$ consists of 88\% and 12\% of the documents respectively. In the second cluster, the words ``card", ``video" and ``driver" occur with relatively high probabilities. It is interpreted as a topic, i.e., ``video-card-driver". The words occur with very low probabilities in the first cluster.  It is regarded as a background topic. Similarly, $Z_{15}$ gives us the topic ``windows-dos" and it consists of 15\% of the documents. The topic given by $Z_{22}$ involves many words related to computers and hence can be simply understood as a topic about computers.

Latent variables at low levels of the hierarchy capture ``short-range" word co-occurrences and hence they give topics that are relatively more specific in  meaning. Latent variables at high levels of the hierarchy capture ``long-range" word co-occurrences and hence they give topics that are relatively more general in meaning. Hence the model gives us a hierarchy of topics, part of which is listed below. HLTA is therefore considered a tool for {\em hierarchical topic detection}.

{\small
\begin{quote}
\begin{tabbing}
\= xxx \= xxx \kill
\> $Z_{22}$: windows card graphics video dos \\
\> \>  $Z_{14}$: card video driver \\
\> \> $Z_{15}$:  windows  dos \\
\> \> $Z_{16}$: graphics display image \\
\> \> $Z_{17}$: computer science
\end{tabbing}
\end{quote}
}

To build a hierarchical model, HLTA first learns a model similar to the one shown in Figure 4 (b). It is a {\em flat LTM} in the sense that every latent variable is directly connected to at least one observed variable. Next, HLTA  converts the latent variables in the flat model ($Z_{11}$, $Z_{12}$, \ldots, $Z_{111}$) into observed variables via data completion, and learns a flat model for them (c). Then, the second flat model is stacked on top of the first one to get the hierarchical model. In general, the process is repeated multiple times to get multiple layers of latent variables.

\cite{liu14hierarchical} use the BI algorithm to learn flat models, which does not scale up.
\cite{chen2015progressive} improves HLTA by using {\em progressive EM (PEM)} to estimate parameters of the intermediate models. The idea is to estimate the parameters in steps and, in each step, EM is run on a submodel that involves only 3 or 4 observed variables. PEM is efficient because a dataset, when projected onto 3 or 4 binary variables, consists of only 8 or 16 distinct cases no matter how large it is.

Topic detection has been one of the most active research areas in machine learning. Nested hierarchical Dirichlet process (nHDP)  \citep{paisley2015nested} is the state-of-the-art method for hierarchical topic detection. Empirical results reported in \citep{liu14hierarchical,chen2015progressive}  show that HLTA significantly outperforms nHDP in terms of topic quality as measured by the topic coherence score proposed by \cite{mimno2011optimizing}, and HLTA finds more meaningful topic hierarchies. Figure 5 shows a part of the topic hierarchy that HLTA obtains from all papers published at AAAI/IJCAI from 2000 to 2015.  It is clearly meaningful.
Interested readers can also browse and compare the topic hierarchies obtained by HLTA and nHDP from 300,000 New York Times articles  at {\small {\tt http://home.cse.ust.hk/\~{ } lzhang/topic/ijcai2016/}}.

\begin{figure}[t]
\includegraphics[width=8.5cm]{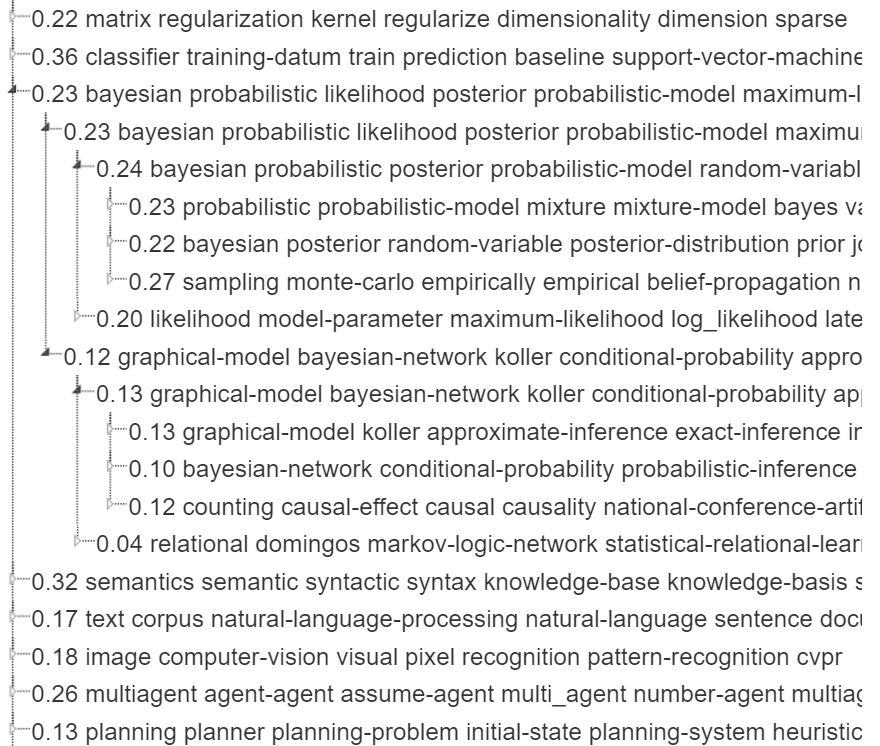}
\vspace{-0.5cm}
\caption{\small  A part of the topic hierarchy  obtained by HLTA  from AAAI/IJCAI (2000 -- 2015) papers.}
\vspace{-0.5cm}
\end{figure}

\section{Deep Probabilistic Modeling}
Deep learning has achieved great successes in recent years. It has produced superior results in a range of applications, including image classification, speech recognition, language translation and so on. Now it might be the time to ask whether it is possible and beneficial to learn structures for deep models.

To learn the structure of a deep model, we need to determine the number of hidden layers and the number of hidden units at each layer. More importantly, we need to determine the connections between neighboring layers. This implies that we need to talk about sparse models where neighboring layers are not fully connected.

Sparseness is desirable and full connectivity is unnecessary. In fact, \cite{NIPS2015_5784} have shown that many weak connections in the fully-connected layers of \emph{Convolutional Neural Networks (CNNs)} \citep{lecun1995convolutional} can be pruned without incurring any accuracy loss.  The convolutional layers of CNNs are sparse, and the fact is considered one of the key factors that lead to the success of CNNs.  Moreover, it is well known that overfitting is a serious problem in deep models. Overfitting is caused not only by excessive amount of hidden units, but also excessive amount of connections. One method to address the problem is dropout \citep{Srivastava:2014:DSW:2627435.2670313}, which randomly drops out units (while keeping full connectivity) during training. Sparseness offers an interesting alternative. It amounts to deterministically drop out connections.

How can one learn sparse deep models? One method is to first learn a fully connected model and then prune weak connections \citep{NIPS2015_5784}. A drawback of this method is that it is computationally wasteful. Moreover, it does not offer a way to determine the number of hidden units. We would like to develop a method that determines the number of hidden units and the connections between units automatically. The key intuition is that a hidden unit should be connected to a group of strongly correlated units at the level below. This idea is used in convolutional layers of CNNs, where a unit is connected to pixels in a small patch of an image. In image analysis, spatial proximity implies strong correlation.

To apply the intuition to applications other than image analysis, we need to identify groups of strongly correlated variables for which latent variables should be introduced. HLTA offers a plausible solution. As explained in the previous section, HLTA first learns a flat LTM. To do so, it  partitions all the variables into groups such that the variables in each group are strongly correlated and the correlations can be properly modeled using a single latent variable \citep{liu14hierarchical,chen2015progressive}. It introduces a latent variable for each group and links up the latent variables to form a flat LTM. Then it converts the latent variables into observed variables via data completion and repeats the process to produce a hierarchy.

 The output of HLTA is a deep tree model with a  layer of  observed variables at the bottom and multiple layers of latent variables on top (see Figure 4 (a)). To obtain a non-tree sparse deep model, we can use the tree model as a skeleton and introduce additional connections to model the residual correlations not captured by the tree.

\cite{chen2016sparse} have  developed and tested the idea in the context of RBMs, which have a single hidden layer and are building blocks of Deep Belief Networks \citep{hinton2006fast} and Deep Boltzmann Machines \citep{salakhutdinov2009deep}.  The target domain is unsupervised text analysis. They have worked out an algorithm for learning what are called Sparse Boltzmann Machines.
The method can determine the number of hidden units and the connections among the units. The models obtained by the method are significantly better, in terms
of held-out likelihood, than RBMs where the hidden and observed
units are fully connected. This is true even when the
number of hidden units in RBMs is optimized by held-out validation.
Moreover, they have demonstrated that Sparse Boltzmann Machines are also more interpretable than RBMs.

\section{Other Applications}

LTMs can be used as a tool for general probabilistic inference over discrete variables. Here one works with a joint distribution over a set of variables and makes inference to compute the posterior distribution of query variables given evidence variables. It is technically challenging because explicit representation of the joint distribution takes space exponential in the number of variables, and inference  takes exponential time.

Bayesian networks \citep{pear1988probabilistic} alleviate the problem by representing the joint distribution in a factorized form. LTMs offer an alternative method. The idea is to build an LTM with the variables in question as observed variables, and make inference with the LTM. LTMs have two attractive properties. On one hand, they are computationally simple to work with because  they are tree structured. On the other hand, they can represent complex relationship among the observed variables. Those two properties are exploited in  \citep{wang08alatent,kaltwang2015latent,yu2016latent} for efficient probabilistic inference in various domains.

LTMs also have a role to play in spectral clustering \citep{poon2012model}.
In spectral clustering \citep{von2007tutorial}, one defines a similarity
matrix for a collection of data points, transforms
the matrix to get a Laplacian matrix,
finds the eigenvectors of the Laplacian
matrix, and obtains a partition of the data
using the leading eigenvectors. The last step
is sometimes referred to as {\em rounding}.

Rounding amounts to clustering the data points using the eigenvectors as features.What is unique about the problem is that one needs to determine how many leading eigenvectors to use. To solve the problem using LTMs,  \cite{poon2012model} binarize the eigenvectors and build a collection of LTMs. Each LTM is built in two steps. In the first step, an LCM is constructed using the first $k$ leading vectors and a partition of data is obtained by LCA. In the second step, subsequent vectors and latent variables are added to the model. The construction of the model is motivated by some theoretical results about the ideal case where between cluster similarity is 0. According to the results, the LTM should fit the data the best when the choice of $k$ is optimal. The problem of choosing among different $k$ is hence turned into the problem of choose among different LTMs.

Unlike alternative methods,  this LTM-based method does not require
 the number of clusters equal the number of leading eigenvectors included, and determines both the number of eigenvectors and the number of clusters automatically.
Empirical  results
show that it outperforms the alternative methods. It works correctly
in the ideal case  and degrades gracefully as one
moves away from the ideal case, which is a desirable behavior for spectral clustering methods.

\section{Conclusions}

Latent tree analysis is a novel tool for correlation modeling. It have been shown to be useful in unidimensional clustering, multidimensional clustering, hierarchical topic detection, deep probabilistic modeling, probabilistic inference and spectral clustering. It is potentially useful also in other areas because modeling correlations among variables is a fundamental task in data analysis.

\section{ Acknowledgments}
Research on this article was
supported by Hong Kong Research Grants Council under grants 16202515 and 16212516, and the Education University of Hong
Kong under project RG90/2014-2015R. We thank all our collaborators, particularly Tao Chen, Yi Wang, Tengfei Liu, Raphael Mourad, April H. Liu, Chen Fu,  Peixian Chen, Zhourong Chen.

{
\small
\bibliographystyle{aaai}
\bibliography{paper}
}

\end{document}